\documentclass[runningheads]{llncs}
\usepackage{graphicx}
\usepackage{amsmath,amssymb} 
\usepackage{color}

\def\onedot{.}
\def\eg{\emph{e.g}\onedot} 
\def\ie{\emph{i.e}\onedot} 
 
\def\etc{\emph{etc}\onedot} 
 
\def\TITLE{Deep Disentangled Semantic Explorative Extreme Super-Resolution}
\def\SHORTTITLE{DeepSEE}
\def\etal{\emph{et al.}}
\makeatother
\newcommand{\fref}[1]{Fig.~\ref{#1}}
\newcommand{\tref}[1]{Table~\ref{#1}}
\newcommand{\eref}[1]{Equation~\ref{#1}}
\newcommand{\sref}[1]{Section~\ref{#1}}
\newcommand{\aref}[1]{the supplementary materials}
\newcommand{\ETAL}[1]{~\etal~\cite{#1}}

\newcommand{\deepsee}{\textit{DeepSEE}}

\begin{document}
\pagestyle{headings}
\mainmatter

\def\ACCV20SubNumber{499}  

\title{\SHORTTITLE: \TITLE} 
\titlerunning{DeepSEE}

\authorrunning{M. C. B\"uhler \etal}

\author{Marcel C. B\"uhler\orcidID{0000-0001-8104-9313}
\and
Andr\'es Romero\orcidID{0000-0002-7118-5175}
\and
Radu Timofte\orcidID{0000-0002-1478-0402}}

\institute{Computer Vision Lab, ETH Z\"urich, Switzerland \email{\{buehlmar,roandres,timofter\}@ethz.ch}}

\maketitle


\begin{abstract}
Super-resolution (SR) is by definition ill-posed. There are infinitely many plausible high-resolution variants for a given low-resolution natural image. Most of the current literature aims at a single deterministic solution of either high reconstruction fidelity or photo-realistic perceptual quality. In this work, we propose an explorative facial super-resolution framework, DeepSEE, for Deep disentangled Semantic Explorative Extreme super-resolution. To the best of our knowledge, DeepSEE is the first method to leverage semantic maps for explorative super-resolution. In particular, it provides control of the semantic regions, their disentangled appearance and it allows a broad range of image manipulations.
We validate DeepSEE on faces, for up to $32\times$ magnification and exploration of the space of super-resolution. 
Our code and models are available at: \url{https://mcbuehler.github.io/DeepSEE/}.
\keywords{
explorative super-resolution, face hallucination, stochastic super-resolution, extreme super-resolution, disentanglement, perceptual super-resolution, generative modeling, disentanglement}
\end{abstract}

\section{Introduction}

\begin{figure}[t]
\centering
\includegraphics[width=\textwidth]{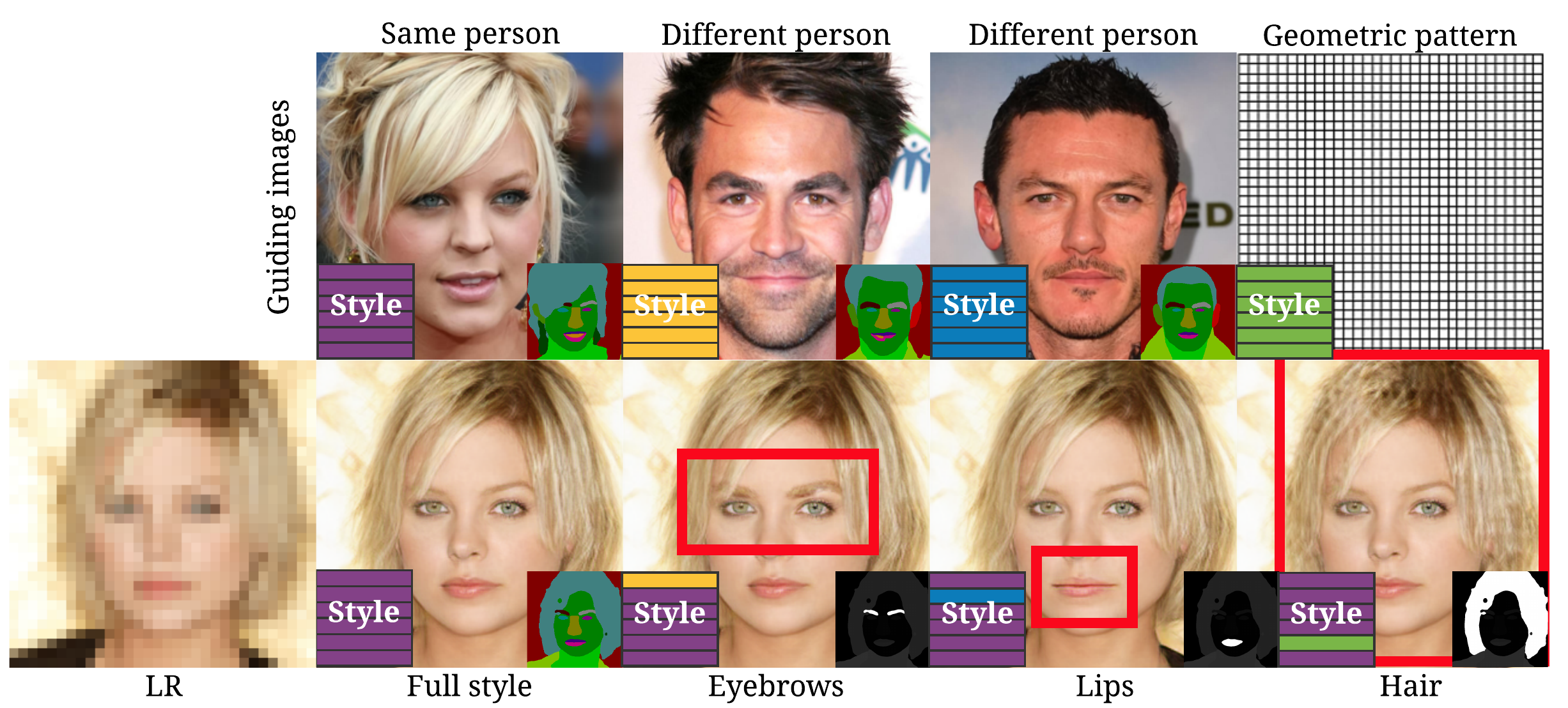}
\caption{\textbf{Upscaling and Manipulations with Disentangled Style Injection.} The bottom row shows the low-resolution input and four high-resolution variants; the top row displays the guiding images. Our model, \deepsee{}, can apply the \textit{full style} matrix from an image of the same person, and alter it with styles from guiding images. We learn $19$ semantic regions, such as \textit{eyebrows}, \textit{lips}, \textit{hair}, etc. \deepsee{} also allows style extraction from \textit{geometric patterns}, sampling in the solution space (\fref{fig:whichone}), style interpolation (\fref{fig:interpolations}), semantic manipulations (\fref{fig:manip_semantics}), and upscaling to extreme magnification factors (\fref{fig:whichone} and \fref{fig:qualitativeextreme}).}
\label{fig:manipulationsref}
\end{figure}

\begin{figure}[t]
\centering
\includegraphics[width=\textwidth]{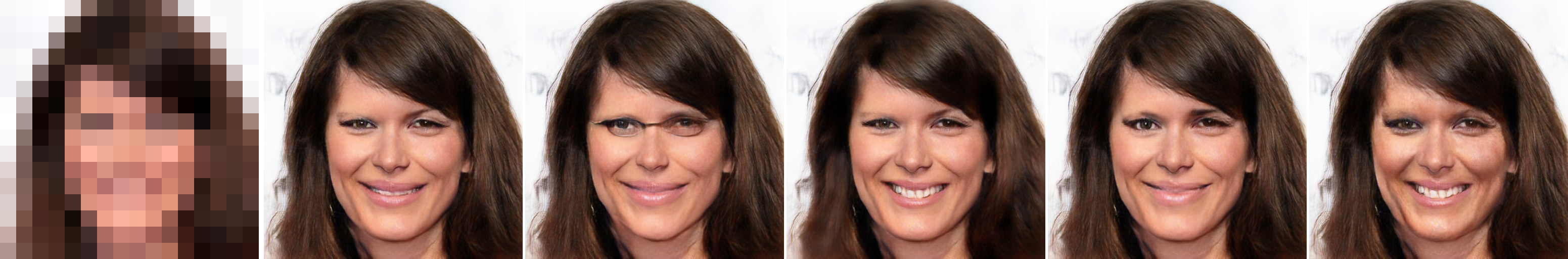}
\caption{\textbf{Multiple Potential Solutions for a Single Input.} We upscale with factor $32\times$ to different high-resolution variants. \textit{Which one would be the correct solution?}}
\label{fig:whichone}
\end{figure}

In super-resolution (SR), we learn a mapping $G_\Theta$ from a low-resolution (LR) image $x_{lr}$ to a higher-resolution (HR) image $\hat x_{hr}$:
\begin{align}
    \hat x_{hr} = G_\Theta(x_{lr}).
\end{align}

Simple methods, like bilinear, bicubic or nearest-neighbour, do not restore high-frequency content or details---their output looks unrealistic. Most modern super-resolution methods rely on neural networks to learn a more complex mapping. Typical upscaling factors are $4\times$ to $8\times$~\cite{timofte2017ntire,blau20182018,cai2019ntire,wang2018recoveringsftgan,yu2018superverylow,li2019deep,kim2019progressive-face-sr,lee2018attribute,chen2018fsrnet,yu2018facemultitask,liu2015faceattributesceleba}; generating $4^2$, respectively $8^2$ pixels for one input pixel. Very recent works upscale up to $16\times$~\cite{shang2020perceptual,gu2019aim,zhang2020ntire} and $64\times$~\cite{PULSE_CVPR_2020}. 

The mapping between the low- and the high-resolution domain is not well defined. There exist multiple (similar) high-resolution images that would downscale to the same low-resolution image. This is why super-resolution is an \textit{ill-posed inverse problem}. Yet, most modern methods assume a ground truth; and learn to generate a single result for a given input~\cite{ledig2017photosrgan,wang2018esrgan,wang2018recoveringsftgan,yu2018superverylow,kim2019progressive-face-sr,lee2018attribute,chen2018fsrnet,yu2018facemultitask,liu2015faceattributesceleba}.

 To add more guidance, some methods leverage additional information to tackle the super-resolution problem. This can include image attributes~\cite{yu2018superverylow,lee2018attribute,liu2015faceattributesceleba,li2019deep}, reference images~\cite{li2018learninggfrnet,dogan2019exemplargwainet} and/or a guidance by facial landmark~\cite{chen2018fsrnet,kim2019progressive-face-sr,yu2018facemultitask}. Still, neither of those approaches allows to produce more than very few variants for a given input. Ideally, we could generate an infinite number of potentially valid solutions an pick the one that suits our purpose best. 

Our proposed method, \deepsee{}, is capable of generating a large number of high-resolution candidates for a low-resolution face image. The outputs differ in both appearance and shape, but they are overall consistent with the low-resolution input. Our method learns a one-to-many mapping from a low-frequency input to a disentangled manifold of potential solutions with the same low-frequencies, but diverse high-frequencies. For inference, a user can specifically tweak the shape and appearance of individual semantic regions to achieve the desired result.
\deepsee{} allows to sample randomly varied solutions (\fref{fig:whichone}), interpolate between solution variants (\fref{fig:interpolations}), control high-frequency details via a guiding image (\fref{fig:manipulationsref}), and manipulate pre-defined semantic regions (\fref{fig:manip_semantics}). In addition, we go beyond common upscaling factors and magnify up to $32\times$.

\subsection{Contributions}

\begin{itemize}
    \item[i)] We introduce \deepsee{}, a novel face hallucination framework for \textbf{Deep} disentangled \textbf{S}emantic \textbf{E}xplorative \textbf{E}xtreme super-resolution.
    
    \item[ii)] We tackle the ill-posed super-resolution problem in an \textbf{explorative} approach based on \textbf{semantic maps}. \deepsee{} is able to sample and manipulate the solution space to produce an infinite number of high-resolution faces for a single low-resolution input.
    
    \item[iii)] \deepsee{} gives control over both shape and appearance. A user can tweak the \textbf{disentangled} \textbf{semantic} regions individually.
    
    \item[iv)] We super-resolve to the \textbf{extreme}, with upscaling factors up to $32 \times$.
\end{itemize}

\section{Related Work}

\subsection{Fidelity vs. Perceptual Quality in Super-resolution}
Single image super-resolution assumes the availability of a low-resolution image carrying low-frequency information---basic colors and shapes---and aims to restore the high-frequencies---sharp contrasts and details. The output is a high-resolution image that is consistent with the low-frequency input image.

Traditional super-resolution methods focused on \textbf{\textit{fidelity}}: low distortion to a high-resolution ground truth image. These methods based on edge~\cite{fattal2007imageedge,sun2008imagegradientprofile} and image statistics~\cite{aly2005image,zhang2010non} and relied on traditional supervised machine learning algorithms: support-vector regression~\cite{ni2007image}, graphical models~\cite{wang2005patch}, Gaussian process regression~\cite{he2011single}, sparse coding~\cite{yang2010image} or piece-wise linear regression~\cite{timofte2014a+}.

With the advent of deep learning, the focus shifted to \textbf{\textit{perceptual}} quality: photo-realism as perceived by humans. 
Their results are less blurry and more realistic~\cite{blau20182018}, defining more and more the current main stream research \cite{ledig2017photosrgan,wang2018esrgan,dong2015image,kim2016deeply,timofte2017ntire,timofte2018ntire,cai2019ntire,zhang2020ntire}.

\subsubsection{Evaluation.}
\label{sec:relatedwork_evaluation}
\label{sec:sreval}
Traditional evaluation metrics in super-resolution are \textit{Peak Signal-to-Noise Ratio} (PSNR) or \textit{Structural Similarity Index}~\cite{wang2004imagessim} (SSIM). However, these \textit{fidelity} metrics are simple functions that measure the distortion to reference images and correlate poorly with the human visual response of the output~\cite{ledig2017photosrgan,blau20182018,wang2018esrgan,wang2018recoveringsftgan}. A high PSNR or SSIM does not guarantee a perceptually good looking output~\cite{blau2018perception}.
Alternative metrics evaluate perceptual quality, namely the \textit{Learned Perceptual Image Patch Similarity} (LPIPS)~\cite{zhang2018perceptuallpips} and the \textit{Fr\'echet Inception Distance} (FID)~\cite{heusel2017ttur}. In this work, we emphasize our validation on high visual quality as in~\cite{ledig2017photosrgan,wang2018esrgan,sajjadi2017enhancenet,wang2018recoveringsftgan}, exploration of the solution space~\cite{bahat2019explorable} and extreme super-resolution~\cite{shang2020perceptual,gu2019aim,zhang2020ntire}.

\subsection{Perceptual Super-resolution}
Super-resolution methods with focus on high fidelity tend to generate blurry images~\cite{blau20182018}. In contrast, perceptual super-resolution targets photo-realism. Training perceptual models typically includes perceptual losses~\cite{johnson2016perceptual,simonyan2014veryvgg}, or Generative Adversarial Networks (GAN)~\cite{ledig2017photosrgan,blau20182018,wang2018esrgan,bulat2018superfan,wang2018recoveringsftgan}.

Generative Adversarial Networks~\cite{goodfellow2014generative} (GAN) have become increasingly popular in image generation~\cite{karras2019style,karras2019analyzing,spade,choi2018stargan,romero2019smit}. The underlying technique is to alternately train two neural networks---a generator and a discriminator---with contrary objectives, playing a MiniMax game. While the discriminator aims to correctly classify images as real or fake, the generator learns to produce photo-realistic images fooling the discriminator. 

 A seminal GAN-based work for perceptual super-resolution, SRGAN~\cite{ledig2017photosrgan}, employed a residual network~\cite{he2016deep} for the generator and relied on a discriminator~\cite{goodfellow2014generative} for realism. A combination of additional losses encourage reconstruction/fidelity and texture/content. ESRGAN~\cite{wang2018esrgan} further improved upon SRGAN by tweaking its architecture and loss functions. 
 
  In this work, we propose a \textit{GAN-based perceptual} super-resolution method.

\subsection{Explorative Super-resolution}
One severe shortcoming of existing approaches is that they consider super-resolution as a \textit{1:1} problem: A low-resolution image maps to a single high-resolution output~\cite{wang2018recoveringsftgan,yu2018superverylow,kim2019progressive-face-sr,lee2018attribute,chen2018fsrnet,yu2018facemultitask,liu2015faceattributesceleba,ledig2017photosrgan,wang2018esrgan}. In reality, however, an infinite number of consistent solutions would exist for a given low-frequency input. Super-resolution is by definition an \textit{ill-posed inverse problem}. Downscaling many (similar) high-resolution variants would yield the same low-resolution image~\cite{cai2019ntire,REN20132408,haris2018deep,li2018sr,lugmayr2020srflow,ravishankar2011sr,PULSE_CVPR_2020}. 
In our work, we regard super-resolution as a \textit{1:n} problem: A low-resolution image maps to many consistent high-resolution variants. 

In a concurrent work, Bahat\ETAL{bahat2019explorable} suggest an editing tool with which a user can manually manipulate the super-resolution output.
Their manipulations include adjusting the variance or periodicity for textures, reducing brightness, or brightening eyes in faces. 
Two recent works leverage normalizing flows~\cite{dinh2016density,kingma2018glow} for non-deterministic super-resolution~\cite{lugmayr2020srflow,xiao2020invertible}.
In our work, we allow to freely walk a latent style space and manipulate semantic masks to explore even more solutions. 

To the best of our knowledge, \cite{bahat2019explorable} and ours are the first works targeting \textit{semantically controllable} explorative super-resolution; and \deepsee{} is the first method that achieves explorative super-resolution using semantically-guided style imposition.

\subsection{Domain-specific Super-resolution}
Typical domain-specific applications include super-resolution of faces~\cite{yu2018superverylow,kim2019progressive-face-sr,lee2018attribute,chen2018fsrnet,yu2018facemultitask,liu2015faceattributesceleba}, outdoor scenes~\cite{wang2018recoveringsftgan} or depth maps~\cite{wang2019deepsurvey,riegler2016atgv,hui2016depth,song2016deep}. Applying super-resolution in a constraint domain allows to leverage prior knowledge and additional guidance, like enforcing characteristics via attributes or identity annotations~\cite{lee2018attribute,yu2018superverylow,liu2015faceattributesceleba,li2019deep}, facial landmarks~\cite{chen2018fsrnet,kim2019progressive-face-sr,yu2018facemultitask}, guiding images~\cite{li2018learninggfrnet,dogan2019exemplargwainet}, or semantic maps~\cite{timofte2016semantic,wang2018recoveringsftgan}.

In this work, we focus on super-resolution for faces, namely \textit{face hallucination}. Despite the important roles of facial keypoints, attributes and identities, they are a high-level supervision that does not allow fine-grained manipulation of the output---oftentimes a desired property.
In contrast to previous works, we use a predicted discrete semantic prior for each region of the face. 

\subsection{Extreme Super-resolution}
Recent extreme super-resolution train on the DIV8K dataset~\cite{gu2019div8k} and target $16\times$ upscaling~\cite{shang2020perceptual,gu2019aim,zhang2020ntire}. A concurrent work~\cite{PULSE_CVPR_2020} searches the latent space of a pre-trained face generation model \cite{karras2019style} to find high-resolution images that match a low-resolution image, when downscaled $64\times$.

\section{\SHORTTITLE}
\label{sec:deepsee}
\begin{figure}[t]
\centering
\includegraphics[width=0.9\textwidth]{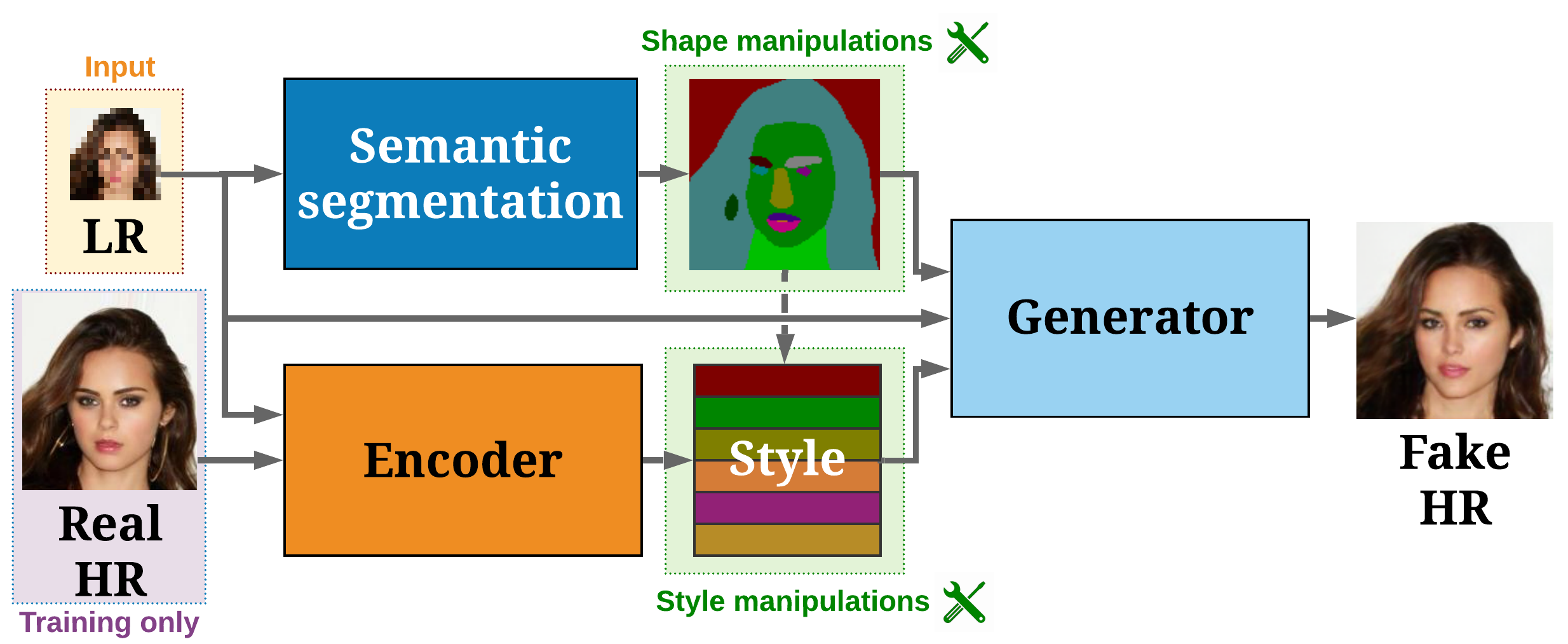}
\caption{\textbf{Overview of Components and Information Flow.} \deepsee{} guides the upscaling with a semantic map extracted from the low-resolution input, and a latent style matrix encoded from a high- or low-resolution image. During inference, a user can tweak the output by manipulating shapes and style codes.
}
\label{fig:archhighlevel}
\end{figure}

\subsection{Problem Formulation}
A low-resolution input ($x_{lr} \in \mathbb{R}^{H_{lr}\times{W_{lr}}\times{3}}$) image acts as a starting point that carries the low-frequency information. A generator ($G_{\Theta}$) upscales this image and hallucinates the high-frequencies yielding the high-resolution image $\hat x_{hr} \in \mathbb{R}^{H_{hr}\times{W_{hr}}\times{3}}$. As a guidance, $G_{\Theta}$ leverages both a high resolution semantic map ($M \in \mathbb{R}^{H_{hr}\times{W_{hr}}\times{N}}$, where $N$ is the number of the semantic regions) and independent styles per region ($S\in\mathbb{R}^{N\times d}$, where $d$ is the style dimensionality). The upscaled image should thus retain the low-frequency information from the low-resolution image. In addition, it should be consistent in terms of the semantic regions and have specific, yet independent styles per region. We formally define our problem as

\begin{equation}
    \hat x_{hr} = G_{\Theta}(x_{lr}, \thinspace M, \thinspace S).
    \label{equation:formulation}
\end{equation}

Remarkably, thanks to the flexible semantic layout, a user is able to control the \textit{appearance} and \textit{shape} of each semantic region through the generation process. This allows to tweak an output until the desired solution has been found.

\subsection{Architecture}
\label{sec:arch}
Following the GAN framework~\cite{goodfellow2014generative}, our method consists of a generator and a discriminator network. In addition, we employ a segmentation network and an encoder for style. Concretely, the segmentation network predicts the semantic mask from a low-resolution image and the encoder produces a disentangled style. \fref{fig:archhighlevel} illustrates our model at a high level and \fref{fig:overallarch} provides a more detailed view. In the following, we describe each component in more detail.

\subsubsection{Style Encoder.}
\label{sec:styleencoder}
The style encoder $E$ extracts $N$ style vectors of size $d$ from an input image and combines them to a style matrix $S \in\mathbb{R}^{N\times d}$. Remarkably, it can extract the style from either a low-resolution image $x_{lr}$ or a high-resolution image $x_{hr}$ and maps the encoded style to the same latent space $\mathcal{S}$. The encoder disentangles the regional styles via the semantic layout $M$. The resulting style matrix serves as guidance for the generator. During inference, a user can sample from the latent style space $\mathcal{S}$ to produce diverse outputs. Please note that the encoder never combines high- and low-resolution inputs; the input is \textit{either} a high-resolution image \textit{or} a low-resolution image.

\begin{figure}[t]
\centering
\includegraphics[width=\textwidth]{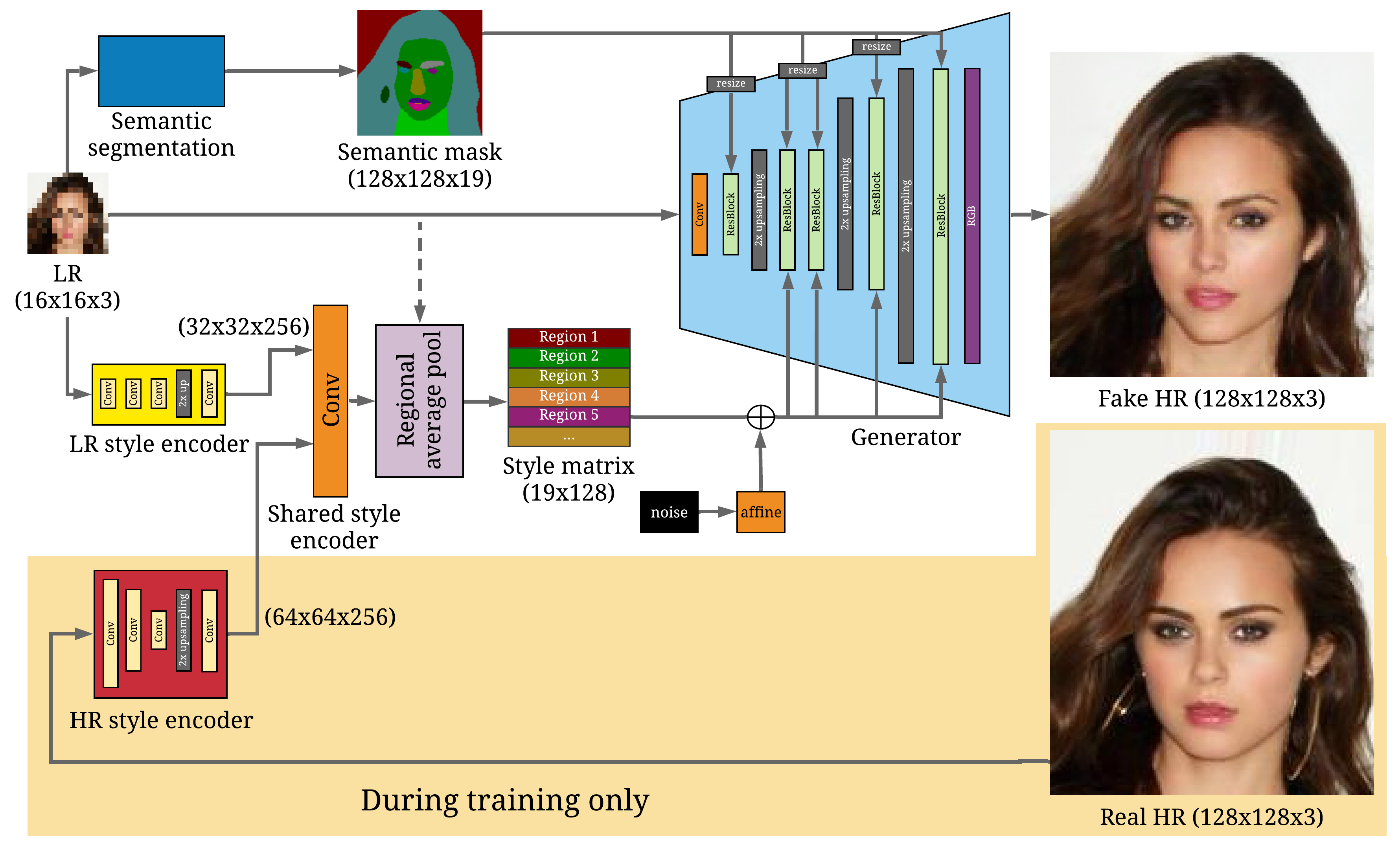}
\caption{\textbf{\deepsee{} Architecture.} Our Generator upscales a low-resolution image (LR) in a series of residual blocks. A predicted semantic mask guides the geometric layout and a style matrix controls the appearance of semantic regions.
The noise added to the style matrix increases the robustness of the model. We describe the style encoding, generator and semantic segmentation in~\sref{sec:arch}.
}
\label{fig:overallarch}
\end{figure}

The style encoder consists of a convolutional neural network  $E_{lr}$ for the low-resolution and a similar convolutional neural network $E_{hr}$ for the high resolution input. Their output is mapped to the same latent style space via a shared layer $E_{Shared}$. \fref{fig:overallarch} illustrates the flow from the inputs to the style matrix. The architecture for the high-resolution input $E_{hr}$ consists of four convolution layers. The input is downsized twice in the intermediate two layers and upsampled again after a bottleneck. Similarly, the low-resolution encoder $E_{lr}$ consists of four convolution layers. It upsamples the feature map once before the shared layer.
The resulting feature map is then passed through the shared convolution layer $E_{Shared}$ and mapped to the range $[-1, 1]$.

Inspired by Zhu\ETAL{zhu2019sean}, as a final step, we collapse the output of the shared style encoder for each semantic mask using regional average pooling. This is an important step to disentangle style codes across semantic regions. We describe the regional average pooling in detail in the supplementary material.

\subsubsection{Generator.}
\label{sec:generator}
Our generator learns a mapping $G_\Theta(x_{lr} | M, S)$, where the model conditions on both a semantic layout $M$ and a style matrix $S$. 
This allows to influence the \textit{appearance}, as well as the \textit{size} and \textit{shape} of each region in the semantic layout.

The semantic layout $M \in \{0, 1\}^{H_{hr} \times W_{hr} \times N}$ consists of one binary mask for each semantic region $\{M_0,\cdots,M_{N-1}\}$. For style, we assume a uniform distribution $S \in {[-1, 1]}^{N \times d}$, where each row in $S$ represents a style vector of size $d$ for one semantic region.

At a high level, the generator is a series of residual blocks with upsampling layers in between. 
Starting from the low-resolution image, it repeatedly doubles the resolution using nearest neighbor interpolation and processes the result in residual blocks.
In the residual blocks, we inject semantic and style information through multiple normalization layers.

For the semantic layout, we use \textit{spatially adaptive normalization} (SPADE)~\cite{spade}. SPADE learns a spatial modulation of the feature maps from a semantic map. For the style, we utilize \textit{semantic region adaptive normalization} in a similar fashion as~\cite{zhu2019sean}. \textit{Semantic region adaptive normalization} is an extension to SPADE, which includes style. Like SPADE, it computes spatial modulation parameters, but also takes into consideration a style matrix computed from a reference image. In our case, we extract the style $S$ from an input image through our style encoder as described in \sref{sec:styleencoder}. For more details, please check the supplementary material.

\subsubsection{Discriminator.}
We use an ensemble of two similar discriminator networks. One operates on the full image, and the another one on the half-scale of the generated image. Each network takes the concatenation of an image with its corresponding semantic layout and predicts the realism of overlapping image patches. The discriminator architecture follows~\cite{spade}. Please refer to the supplementary material for a more detailed description.

\subsubsection{Segmentation Network.}
\label{sec:segmentation}
Our training scheme assumes high-resolution segmentation maps, which in most cases are not available during inference. Therefore, we predict a segmentation map from the low-resolution input image $x_{lr}$. Particularly, we train a segmentation network to learn the mapping $M = Seg(x_{lr})$, where $M \in \{0, 1\}^{H_{hr} \times W_{hr} \times N}$ is a high-resolution semantic map.

\subsection{\deepsee{} Model Variants}
\label{sec:deepseevariants}
We suggest two slightly different variants of our proposed method. The \textit{guided} model learns to super-resolve an image with the help of a high-resolution (HR) reference image. The \textit{independent} model does not require any additional guidance and infers a reference style from the low-resolution image. 

The \textit{guided} model is able to apply characteristics from a reference image. When fed a guiding image from the same person, it extracts the original characteristics (if visible). Alternatively, when feeding an image from a different person, it integrates those aspects (as long as it is consistent with the low-resolution input). \fref{fig:manipulationsref} shows an example, where we first generate an image with the style from the same person and then alter particular regions with styles from other images.
The second (\textit{independent}) model applies to the case where no reference image is available.

The \textit{independent} and the \textit{guided} differ in the way the style matrix $S$ is computed. For the \textit{independent} model, we extract the style from the low-resolution input image: $S = E(x_{lr})$. In contrast, the \textit{guided} model uses a high-resolution reference image $x^{ref}_{hr}$ to compute the style $S = E(x^{ref}_{hr})$. It is worth to mention that for training, paired supervision is not necessary as we only require \textit{one} high resolution picture of a person.

\subsection{Training}
The semantic segmentation network is trained independently from the other networks.

We train the generator, encoder and discriminator end-to-end in an adversarial setting, similar to~\cite{spade,wang2018pix2pixhd}. As a difference, we inject noise at multiple stages of the generator. We list hyper-parameters and training details in the supplementary material. In the following, we describe the loss function and explain the noise injection.

\subsubsection{Loss Function.}

Our loss function is identical to~\cite{spade}. Our discriminator computes an adversarial loss $\mathcal{L}_{adv}$ with feature matching $\mathcal{L}_{feat}$: the L1 distance between the discriminator features for the real and the fake image.
In addition, we employ a perceptual loss $\mathcal{L}_{vgg}$ from a VGG-19 network~\cite{simonyan2014veryvgg}. We define our full loss function in \eref{eq:loss}:

\begin{equation}
    \mathcal{L} = \mathcal{L}_{adv} + \lambda_{feat} \mathcal{L}_{feat} + \lambda_{vgg} \mathcal{L}_{vgg}
    \label{eq:loss}
\end{equation}
We set the loss weights to $\lambda_{feat} = \lambda_{vgg} = 10$; please refer to the supplementary material for more details.

\subsubsection{Injection of Noise.}
\label{sec:injectionnoise}
 After encoding the style to a style matrix $S$, we add uniformly distributed noise. We define the noisy style matrix $S'$ as $S' = S + U$, where $U_{ij} \sim Uniform(-\delta, +\delta)$. We empirically choose $\delta$ based on the model variant.

\section{Experimental Framework}
\label{sec:experimental_framework}
\subsection{Datasets}
We train and evaluate our method on face images from CelebAMask-HQ~\cite{karras2017progressive,CelebAMask-HQ} and CelebA~\cite{liu2015faceattributesceleba}. We use the official training splits for developing and training and test on the provided test splits. All low-resolution images (serving as inputs) are computed via bicubic downsampling.
The supplementary material shows qualitative results on the Flickr-Faces-HQ Dataset~\cite{karras2019style} and on outdoor scenes from ADE20K~\cite{zhou2017sceneade20k,wang2018recoveringsftgan}.

\subsection{Semantic Segmentation}
We train a segmentation network~\cite{chen2017deeplab,chen2017rethinkingdeeplabv3} on images from CelebAMask-HQ~\cite{CelebAMask-HQ,karras2017progressive}. The network learns to predict a high-resolution segmentation map with $19$ semantic regions from a low-resolution image. As a model, we choose DeepLab V3+~\cite{chen2017deeplab,chen2017rethinkingdeeplabv3,chen2018encoderdeeplab} with DRN~\cite{Yu2016multiscale,Yu2017drn} as the backbone.

\subsection{Baseline and Evaluation Metrics}

We establish a baseline via bicubic interpolation; we first downsample an image to a low-resolution and then upsample it back to the high resolution.

We compute the traditional super-resolution metrics \textit{peak signal-to-noise ratio} (PSNR), \textit{structural similarity index} (SSIM)~\cite{wang2004imagessim}
and the perceptual metrics \textit{Fr\'echet Inception Distance} (FID)~\cite{heusel2017ttur} and \textit{Learned Perceptual Image Patch Similarity} (LPIPS)~\cite{zhang2018perceptuallpips}. 
Our method focuses on generating results of high perceptual quality, measured by LPIPS and FID. 
PSNR and SSIM are frequently used, however, they are known not to correlate very well with perceptual quality~\cite{zhang2018perceptuallpips}. However, we still list SSIM scores for completion and report PSNR in the supplementary material.

\section{Discussion}

\begin{table}[t]
\begin{center}
\caption{
We compare with related work for $8\times$ upscaling on high-resolution images of size $128\times 128$ (CelebA~\cite{liu2015faceattributesceleba}) and $256 \times 256$ (CelebAMask-HQ~\cite{CelebAMask-HQ,kim2019progressive-face-sr}). We compute all metrics on the official test sets and list quantitative metrics for related work where checkpoints are available. Both our \deepsee{} variants outperform the other methods on the perceptual metrics (LPIPS~\cite{zhang2018perceptuallpips} and FID~\cite{heusel2017ttur}). For qualitative results, please look at \fref{fig:qualitativeguided} and the supplementary material.
}
\setlength{\tabcolsep}{3pt}
\label{tbl:quantresultsall} 

\begin{tabular}{l|rrr | rrr}

 & \multicolumn{3}{c}{a) $128\times 128$} & \multicolumn{3}{c}{b) $256\times 256$} \\
\hline\noalign{\smallskip}
Method & SSIM  $\uparrow$& LPIPS $\downarrow$ & FID $\downarrow$ & SSIM  $\uparrow$& LPIPS $\downarrow$ & FID $\downarrow$\\
\noalign{\smallskip}
\hline
\noalign{\smallskip}
Bicubic 
    & $0.5917$ 
    & $0.5625$ 
    & 159.60 

    & $0.6635$ 
    & $0.5443$ 
    & $125.15$ \\
    
FSRNet (MSE)~\cite{chen2018fsrnet}
            &$0.5647$ 
            & $0.2885$  
            & $54.48$ &-&-&- \\
FSRNet (GAN)~\cite{chen2018fsrnet}
            & $0.5403$  
            & $0.2304$ 
            & $55.62$  &-&-&-\\
Kim\ETAL{kim2019progressive-face-sr} 
    & \textbf{0.6634}
    & $0.1175$ 
    & 11.41 &-&-&-\\
    
GFRNet~\cite{li2018learninggfrnet}  &-&-&-
    & $0.6726$ 
    & $0.3472$ 
    & $55.22$ \\

GWAInet~\cite{dogan2019exemplargwainet} &-&-&-
    & $0.6834$
    & $0.1832$ 
    & $28.79$ \\
\hline 
ours (indep.) 
    & $0.6631$  
    & \textbf{0.1063}  
    & 13.84
    
    & $0.6770$ 
    & $0.1691$ 
    & 22.97\\
    
ours (guided) 
    & $0.6628$ 
    & $0.1071$ 
    & \textbf{11.25}
    
    & \textbf{0.6887} 
    & \textbf{0.1519}
    &\textbf{22.02} \\ 
\end{tabular}
\end{center}
\end{table}

 We validate our method on two different setups. First, we compare with state-of-the-art methods in face hallucination and provide both quantitative and qualitative results. Second, we show results for extreme and explorative super-resolution by applying numerous manipulations for $32 \times$ upscaling.
 
\subsection{Comparison to Face Hallucination Methods} 
To the best of our knowledge, our method is the first face hallucination model based on discrete semantic masks. We compare with \textit{(i)} models that use reference images~\cite{li2018learninggfrnet,dogan2019exemplargwainet} and \textit{(ii)} models guided by facial landmarks~\cite{chen2018fsrnet,kim2019progressive-face-sr}. 

\begin{figure}[t]
\centering
\includegraphics[width=0.9\textwidth]{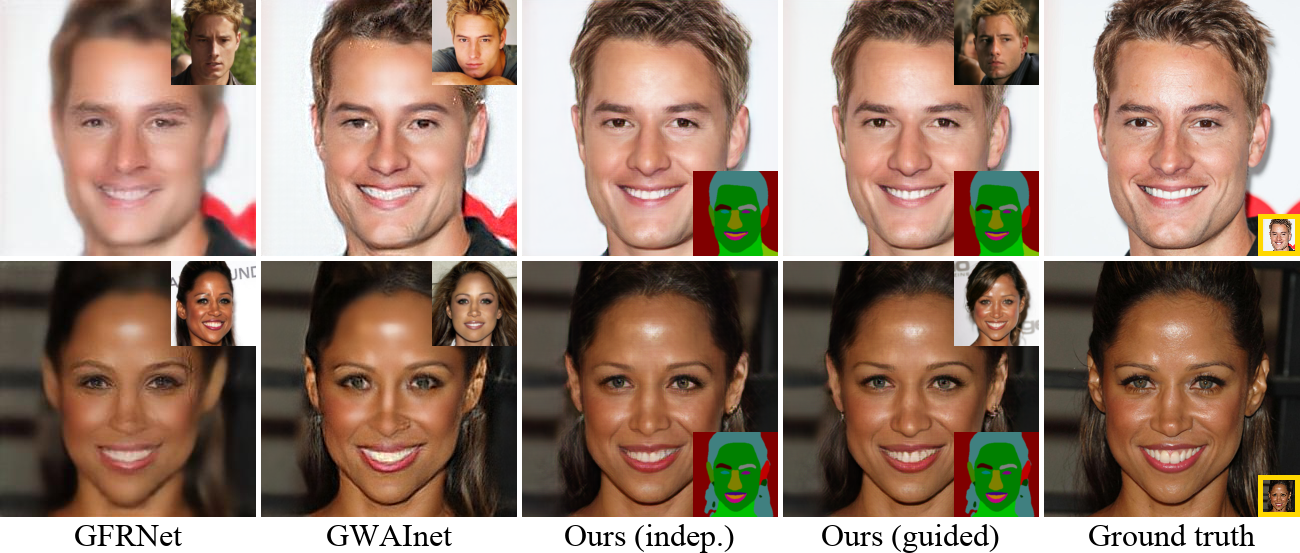}
\caption{\textbf{Comparison to Related Work on 8$\times$ Upscaling}. We compare with our default solutions for the \textit{independent} and \textit{guided} model. The randomly sampled guiding images are on the top right of each image; the bottom right corner shows the predicted semantic mask (if applicable). 
 Our results look less blurry than GFRNet~\cite{li2018learninggfrnet}. Comparing to GWAInet~\cite{dogan2019exemplargwainet}, we observe differences in visual quality for difficult regions, like hair, eyes or teeth. With the additional semantic input, our method can produce more realistic textures. Please zoom in for better viewing.
 }
\label{fig:qualitativeguided}
\end{figure}

For \textit{(i)}, we compare with GFRNet~\cite{li2018learninggfrnet} and GWAInet~\cite{dogan2019exemplargwainet}, both of which leverage an image from the same person to guide the upscaling. Our method achieves the best scores for all metrics in~\tref{tbl:quantresultsall} (b). For perceptual metrics, LPIPS~\cite{zhang2018perceptuallpips} and FID~\cite{heusel2017ttur}, \deepsee{} outperforms the other methods by a considerable margin.
As we depict in~\fref{fig:qualitativeguided}, our proposed method also produces more convincing results, in particular for difficult regions, such as eyes, hair and teeth. We provide more examples in the supplementary material.

For models based on facial landmarks \textit{(ii)}, \tref{tbl:quantresultsall} (a) compares \deepsee{} with FSRNet~\cite{chen2018fsrnet} and Kim\ETAL{kim2019progressive-face-sr}.\footnote{The models from~\cite{chen2018fsrnet,kim2019progressive-face-sr} were trained to generate images of size $128\times128$, so we can evaluate in their setting on CelebA.~\cite{li2018learninggfrnet,dogan2019exemplargwainet} generate larger images ($256\times 256$, whereas CelebA images have size $218 \times 178$), hence we evaluate on CelebAMask-HQ~\cite{karras2017progressive,CelebAMask-HQ}.}
\deepsee{} achieves the highest scores for LPIPS and FID. The supplementary material contains a visual comparison.

\begin{figure}[t]
\centering
\includegraphics[width=\textwidth]{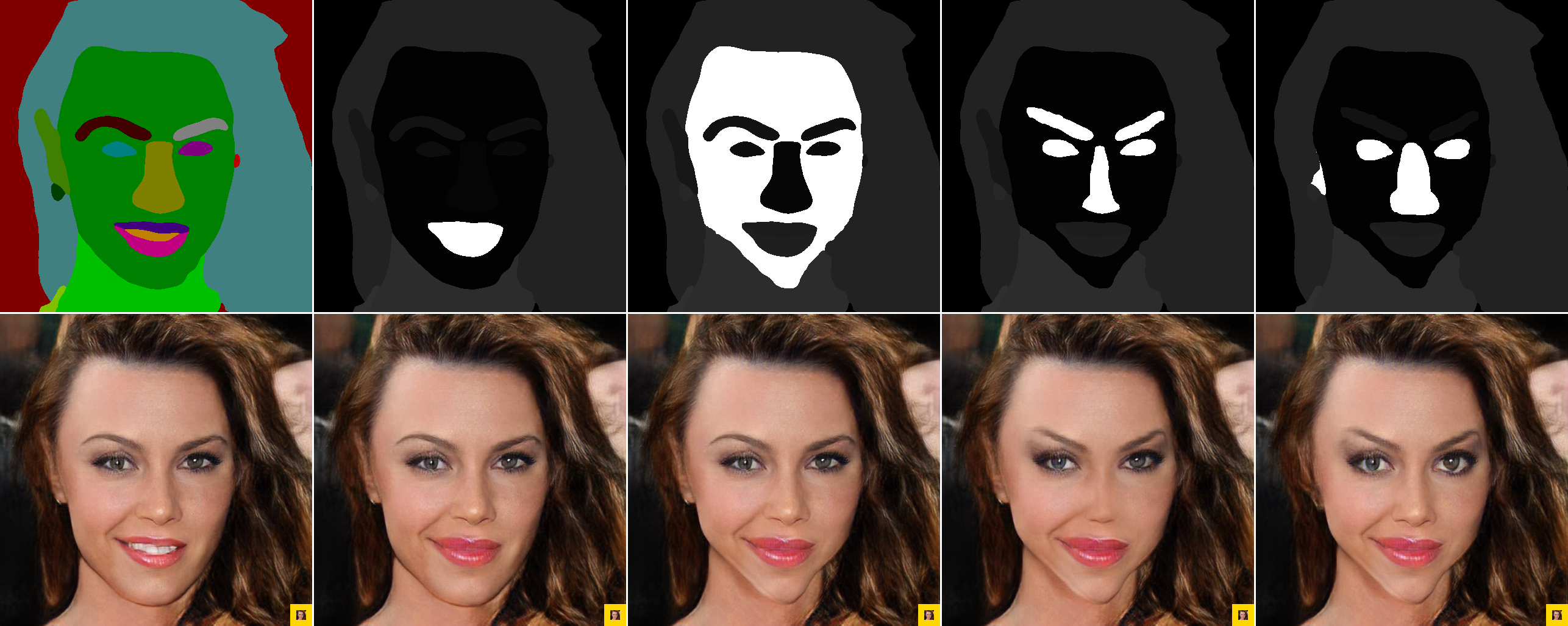}
\caption{\textbf{Manipulating Semantics for $32\times$ Upscaling.} We continuously manipulate the semantic mask and change regional shapes, starting at the default solution (in the first column). In each subsequent column, we highlight the manipulated region and show the resulting image.
}
\label{fig:manip_semantics}
\end{figure}

It is important to note that given the same inputs (\eg{} a low-resolution and a guiding image), all related face hallucination models output a single solution; despite the fact that there would exist multiple valid results. In contrast, our method can generate an infinite number of solutions, and provides the user with fine-grained control over the output. \fref{fig:manipulationsref} and \fref{fig:whichone} show several consistent solutions for a low-resolution image. \deepsee{} can not only extract the overall appearance from a guiding image of the same person, but it can also inject aspects from other people; and even leverage completely different style images, for instance, geometric patterns (\fref{fig:manipulationsref}). We describe \fref{fig:manipulationsref} in more detail in~\sref{sec:manipulations}. In addition, our method allows to manipulate semantics, \ie{} changing the shape, size or content of regions (\fref{fig:manip_semantics}), \eg{} eyeglasses, eyebrows, noses, lips, hair, skin, etc. We provide various additional visualizations in the supplementary material.

\subsection{Manipulations}
\label{sec:manipulations}

\begin{figure}[t]
\centering
\includegraphics[width=\textwidth]{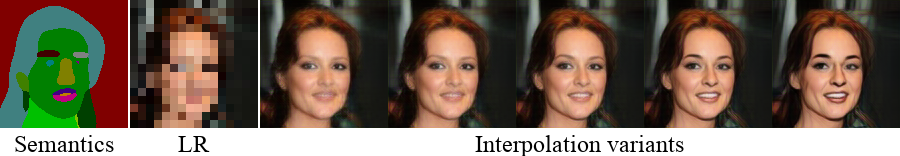}
\caption{\textbf{Interpolation in the Style Latent Space.}  We linearly interpolate between two style matrices, smoothly increasing contrast. Please refer to the supplementary material for more examples.
}
\label{fig:interpolations}
\end{figure}

Our proposed approach is an explorative super-resolution model, which allows a user to tune two main \textit{knobs}---the style matrix and the semantic layout---in order to manipulate the model output. \fref{fig:archhighlevel} shows these knobs in green boxes. 

\subsubsection{Style Manipulations.} The first way to change the output image is to adapt the disentangled style matrix; for instance by adding random noise (supplementary material), by interpolating between style codes (\fref{fig:interpolations}), or by mixing multiple styles (\fref{fig:manipulationsref}).
Going from one style code to another gradually changes the image output. For example, interpolating between style codes can make contrasts slowly disappear, or on the contrary, become more prominent (\fref{fig:interpolations}).

\subsubsection{Semantic Manipulations.} The second tuning knob is the semantic layout. The user can change the size and shape of semantic regions, which causes the generator to adapt the output representation accordingly. \fref{fig:manip_semantics} shows an example where we close the mouth and make the chin more pointy by manipulating the regions for lips and facial skin. Furthermore, we change the shape of eyebrows, reduce the nose and update the stroke of the eyebrows. It is also possible to create hair on a bold head or add/remove eyeglasses (please check the supplementary material). The manipulations should not be too strong, unrealistic or inconsistent with the low-resolution input. Our model is trained with a strong low-resolution prior and hence, only allows relatively subtle shape manipulations.

\subsection{Extreme Super-resolution}

\begin{figure}[t]
\centering
\includegraphics[width=\textwidth]{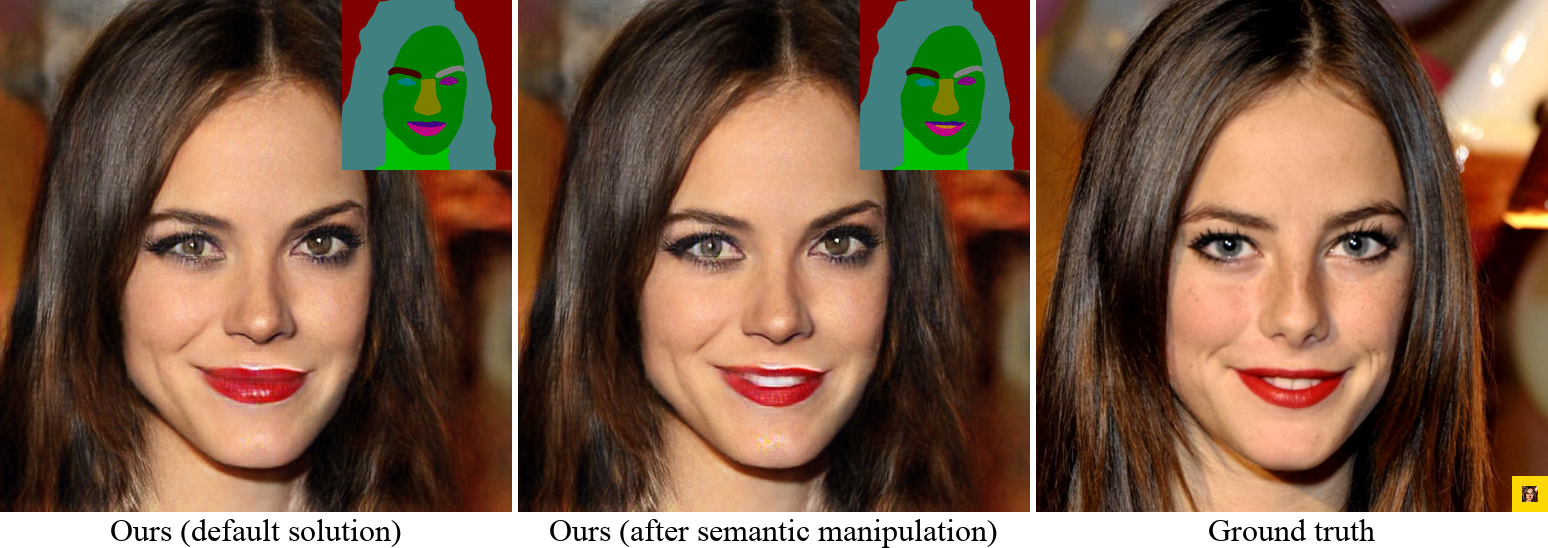}
\caption{\textbf{Extreme Super-resolution.} We show how manipulations can align the model output with an expected outcome. Our default solution shows a closed mouth, while given the ground truth, we would expect a smile. After manipulating the semantic mask, \deepsee{} produces an image that very closely resembles the ground truth.
}
\label{fig:qualitativeextreme}
\end{figure}

While most previous methods apply upscaling factors of $8\times$~\cite{chen2018fsrnet,dogan2019exemplargwainet,li2018learninggfrnet,kim2019progressive-face-sr} or $16\times$~\cite{shang2020perceptual,gu2019aim,zhang2020ntire}, \deepsee{} is capable of going beyond---with upscaling factors of up to $32\times$. 
Instead of reconstructing a single target, \deepsee{} can generate multiple variants in a controlled way and hence, a user is more likely to find an expected outcome. \fref{fig:qualitativeextreme} shows an example where the default solution does not perfectly match the ground truth image. A user can now manipulate the semantic mask and create a second version, which is closer to the ground truth image. This shows the power of explorative super-resolution techniques for extreme upscaling factors.

\section{Ablation Study}
\label{sec:ablation}

\begin{table}[t]
\begin{center}
\caption{\textbf{Ablation Study Results.} We explore the effect of style and semantics. Semantics have the strongest influence on both fidelity (PSNR, SSIM~\cite{wang2004imagessim}) and visual quality (LPIPS~\cite{zhang2018perceptuallpips}, FID~\cite{heusel2017ttur}), but the best results require both semantics and style. Finally, using a high-resolution guiding image (\textit{guided}) from the same person provides an additional point of control to the user compared with the \textit{independent} model.
}
\label{tbl:ablation}
\setlength{\tabcolsep}{3pt}
\begin{tabular}{lcccrrrr}
\hline\noalign{\smallskip}
Name & Semantics & LR-Style & HR-Style &  SSIM  $\uparrow$ & LPIPS $\downarrow$ & FID $\downarrow$ \\
\noalign{\smallskip}
\hline
\noalign{\smallskip}
\textit{Prior-only} & -   & -     &  -    &  0.6168  
    & 0.1233 & 25.11 \\
\textit{LR-style-only}& -   & \checkmark     &  -    & 0.6485
    &0.1103 & 19.29 \\
\textit{HR-style-only}& -   & -     &  \checkmark    & 0.6507
    & 0.1108 & 16.66 \\
\textit{Semantics-only}& \checkmark   & -     &  -    &   0.6543
    & 0.1096 & 12.57 \\
    \hline
\textit{Independent}& \checkmark   & \checkmark     &  -        & \textbf{0.6631} 
    &\textbf{0.1063}  & 13.84\\
\textit{Guided}& \checkmark   & \checkmark     &  \checkmark    &   $0.6628$
    & $0.1071$ 
    & \textbf{11.25}\\

\hline
\end{tabular}
\end{center}
\end{table}

We investigate the influence of \deepsee{}'s main components---semantics and style injection---in an ablation study. \sref{sec:ablation_arch} describes the study setup and we discuss the outcome in \sref{sec:disc_ablation}.

\subsection{Ablation Study Setup}
\label{sec:ablation_arch}

We train four additional models, where we remove the components that inject semantics and/or style. For the first model (\textit{prior-only}), we disable both semantics and style---the model's only conditioning is on the low-resolution input. For the \textit{LR-style-only} and \textit{HR-style-only} models, we do not use the semantic map, but we do condition on the style matrix computed from another low- / high-resolution image of the same person. Lastly, we train a \textit{semantic-only} model that does not inject any style but conditions on semantics. 

All models are trained for 7 epochs, which corresponds to 3 days on a single TITAN Xp GPU, with upscaling factor $8\times$ and batch size 4. We use the CelebA~\cite{liu2015faceattributesceleba} dataset. For details, please check the supplementary material.

\subsection{Ablation Discussion}
\label{sec:disc_ablation}
All performance scores improve when adding either semantics, style or both (\tref{tbl:ablation}). Comparing models with either semantics or style (\textit{LR-style-only} and \textit{HR-style-only} vs. \textit{semantics-only}), the perceptual metrics (LPIPS~\cite{zhang2018perceptuallpips} and FID~\cite{heusel2017ttur}) show better scores when including semantics. Combining both semantic and style yields even better results for both the distortion measures (PSNR and SSIM~\cite{wang2004imagessim}) and the visual metrics (LPIPS~\cite{zhang2018perceptuallpips} and FID~\cite{heusel2017ttur}). The performance between our two suggested model variants (the \textit{independent} model and \textit{guided} model) is very similar for fidelity metrics. In terms of perceptual quality, the \textit{guided} image clearly beats the \textit{independent} in FID. However, we empirically find that the \textit{independent} model is more flexible towards random manipulations of the style matrix. Please refer to the supplementary material for visual examples.

\section{Conclusion}
\label{sec:conclusion}
The super-resolution problem is ill-posed because most high-frequency information is missing and needs to be hallucinated. In this paper, we tackle super-resolution in an explorative approach, \deepsee{}, based on semantic regions and disentangled style codes. 
\deepsee{} allows for fine-grained control of the output, disentangled into region-dependent appearance and shape. Our model goes beyond common upscaling factors and allows to magnify up to $32\times$. Our validation for faces demonstrate results of high perceptual quality.

Interesting directions for further research could be to identify meaningful directions in the latent style space (\eg{} age, gender, illumination, contrast, \etc{}), or to apply \deepsee{} to other domains.
\\
\\
\noindent\textbf{Acknowledgments.}
We would like to thank the Hasler Foundation. This work was partly supported by the ETH Z\"urich Fund (OK), by Huawei, Amazon AWS and Nvidia grants.

\bibliographystyle{splncs}
\bibliography{0499}

\end{document}